\documentclass[10pt,twocolumn,letterpaper]{article}

\usepackage[pagenumbers]{cvpr} 

\usepackage[utf8]{inputenc}

\usepackage{inconsolata}

\usepackage{url}            
\usepackage{booktabs}       
\usepackage{multirow}
\usepackage{bbding} 

\usepackage{times}
\usepackage{latexsym}

\usepackage{amsfonts}       
\usepackage{amsthm,amsmath,amssymb}
\usepackage{bbm,bm}

\usepackage{mathrsfs}
\usepackage{nicefrac}       
\usepackage{microtype}      

\usepackage{graphicx}
\usepackage{float}
\usepackage[countmax]{subfloat}
\usepackage{lipsum}

\usepackage{enumitem}

\usepackage{verbatim}

\usepackage[dvipsnames]{xcolor}

\definecolor{cvprblue}{rgb}{0.21,0.49,0.74}
\usepackage[pagebackref,breaklinks,colorlinks,citecolor=cvprblue]{hyperref}
\usepackage{threeparttable}
\usepackage{bbding}

\usepackage[capitalize]{cleveref}
\crefname{table}{Tab.}{Tab.}
\crefname{figure}{Fig.}{Fig.}
\crefname{section}{Sec.}{Sec.}
\crefname{equation}{Eq.}{Eq.}

\newcommand\Mark[1]{\textsuperscript#1}

\def\paperID{ID} 
\def\confName{CVPR}
\def\confYear{2024}

\title{Multi-scale 2D Temporal Map Diffusion Models for \\Natural Language Video Localization}

\author{Chongzhi Zhang\Mark{1}\Mark{,}\Mark{4}, Mingyuan Zhang\Mark{1}\Mark{,}\Mark{4}, Zhiyang Teng\Mark{2}\Mark{,}\Mark{*}, Jiayi Li\Mark{4}, Xizhou Zhu\Mark{3},\\Lewei Lu\Mark{3}, Ziwei Liu\Mark{1}\Mark{,}\Mark{4}, Aixin Sun\Mark{1}\Mark{,}\Mark{4}\Mark{,}\Mark{\Envelope}
\\
\Mark{1}S-Lab, Nanyang Technological University\quad\Mark{2}ByteDance\quad\Mark{3}SenseTime\\
\Mark{4}School of Computer Science and Engineering, Nanyang Technological University
\\
{\tt\small \{chongzhi001, mingyuan001, jli10\}@e.ntu.edu.sg, chihyangteng@gmail.com,}\\
{\tt\small \{zhuxizhou, luotto\}@sensetime.com, \{axsun, ziwei.liu\}@ntu.edu.sg}
}

\begin{document}
\maketitle

\newcommand\blfootnote[1]{%
\begingroup
\renewcommand\thefootnote{}\footnote{#1}%
\addtocounter{footnote}{-1}%
\endgroup
}
\blfootnote{\Mark{*}Work done while at S-Lab.}
\blfootnote{\Mark{\Envelope}Corresponding author.}

\begin{abstract}
Natural Language Video Localization (NLVL), grounding phrases from natural language descriptions to corresponding video segments, is a complex yet critical task in video understanding. Despite ongoing advancements, many existing solutions lack the capability to globally capture temporal dynamics of the video data. In this study, we present a novel approach to NLVL that aims to address this issue. Our method involves the direct generation of a global 2D temporal map via a conditional denoising diffusion process, based on the input video and language query. The main challenges are the inherent sparsity and discontinuity of a 2D temporal map in devising the diffusion decoder. To address these challenges, we introduce a multi-scale technique and develop an innovative diffusion decoder. Our approach effectively encapsulates the interaction between the query and video data across various time scales. Experiments on the Charades and DiDeMo datasets underscore the potency of our design.
    
\end{abstract}    
\section{Introduction}
\label{sec:intro}

\begin{figure}[t]
\centering
\includegraphics[width=1\linewidth]{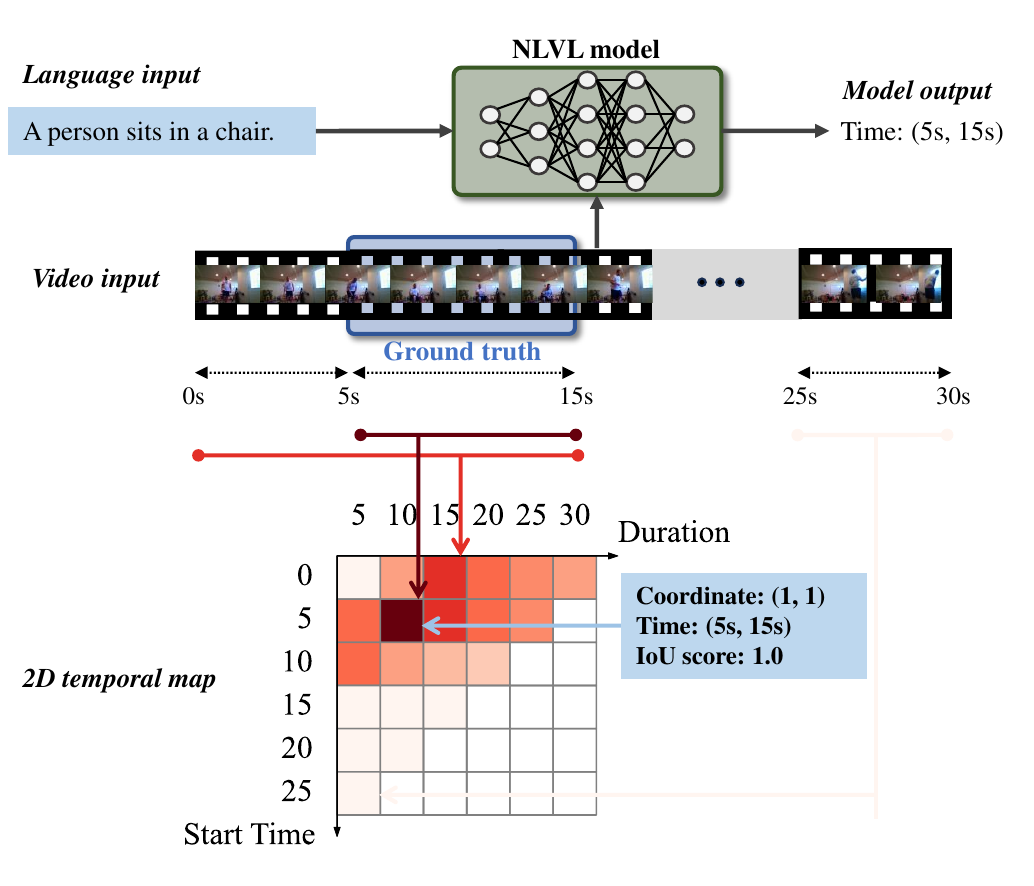} \\
\caption{ 
Illustration of the NLVL task (top part) and 2D temporal map (bottom part). Top: The NLVL model processes a language query and an untrimmed video to locate a temporal moment that semantically corresponds to the query. Bottom: The 2D temporal map plots candidate moments at coordinates $(i, j)$, starting at $i\tau$ and lasting for $(j+1)\tau$; here $\tau=5\text{s}$ is the time scale. The map is displayed as a heatmap, where the values in cells indicate the predicted matching scores between candidate moments and the target moment. Note that the ``ground truth" in the figure is for illustration purpose, and is not available during model inference.}
\label{fig:task_intro}
\end{figure}

Natural Language Video Localization (NLVL) is an essential and significant task in the field of video understanding, boasting a variety of application opportunities. Given an untrimmed video and a query sentence, the NLVL task is to retrieve a temporal moment that semantically corresponds to the query. The top part of \cref{fig:task_intro} provides an example of the NLVL task, where the NLVL model returns the start and end time stamps of a video moment in response to the query ``a person sits in a chair".

Several methods have been proposed to tackle the NLVL task. Regression-based methods~\cite{yuan2019find,lu2019debug,DBLP:conf/aaai/ChenLTXZTL20} directly estimate the temporal timing of target moments by leveraging cross-modal interactions between videos and queries. Span-based methods~\cite{ghosh2019excl,DBLP:conf/acl/ZhangSJZ20,lei2020tvr,zhang2021natural} approach NLVL as a span prediction task, predicting the likelihood of each video segment being the start and end of the target moment. Proposal-based methods~\cite{DBLP:conf/aaai/Xu0PSSS19,xiao2021boundary,xiao2021natural} generate candidate moments, then select the most relevant proposal for a given query. Among the latter, 2D-Map methods~\cite{wang2021structured,DBLP:journals/pami/ZhangPFLL22} utilize a two-dimensional map to represent temporal relationships among proposals. In a 2D temporal map with $\tau$ as the time unit, the coordinates $(i, j)$ indicate a candidate moment starting at $i\tau$ and lasting for $(j+1)\tau$, as shown in the bottom part of \cref{fig:task_intro}. This representation captures all feasible proposals of varying lengths and maintains their adjacent relationships. By including data from neighboring moments, the model is able to gain a comprehensive view of video content, enhancing its understanding and prediction of temporal relationships.

Unlike traditional approaches that treat NLVL as an understanding problem, we frame it as a generative task, using a multi-scale 2D temporal map diffusion model. That is, we adopt a novel perspective on the NLVL task, leveraging the capabilities of diffusion models, a class of generative models renowned for their success in image and text generation. The diffusion model operates on the principle of systematically perturbing data through a forward diffusion process and subsequently recovering the original data via a learned reverse diffusion process. Accordingly, our approach reimagines the NLVL task as the generation of a 2D temporal map, conditioned on both video and language inputs.

Here, the fundamental differences between generative and understanding tasks present unique challenges in designing an effective diffusion decoder. To this end, we have developed a specialized condition-injected decoder, featuring a meticulously crafted architecture that seamlessly integrates conditions and temporal information. This tailored approach ensures the efficient transfer of the diffusion process to the NLVL task. Additionally, we employ multi-scale techniques to enhance performance, facilitating the interaction between query and video data across various temporal scales. Experimental results on the Charades-STA and DiDeMo datasets validate the effectiveness of our innovative design. Our contributions are summarized as follows:

\begin{itemize}
    \item  We propose a novel interpretation of the NLVL task as a diffusion generation problem, specifically focusing on generating multi-scale 2D temporal maps. These maps are conditioned on both video and language inputs. This exploration presents a fresh perspective on applying diffusion models in multimodal understanding tasks.
    \item We identify a fundamental challenge in directly applying successful diffusion models from generation tasks to NLVL. We demonstrate the infeasibility of transplanting existing diffusion models and highlight the inherent differences between understanding tasks and generative tasks. Then, we design a customized diffusion decoder that incorporates conditions and time information in a novel way specific to NLVL.
\end{itemize}

\section{Related Work}

\subsection{The NLVL Task}

Existing frameworks to solve NLVL can be broadly classified into proposal-based and proposal-free frameworks. Within the proposal-free framework, common approaches include regression and span-based methods. Regression-based methods~\cite{yuan2019find,lu2019debug,DBLP:conf/aaai/ChenLTXZTL20} address video localization by learning cross-modal interactions between video and query, and directly estimate the temporal time of target moment. Span-based methods~\cite{ghosh2019excl,DBLP:conf/acl/ZhangSJZ20,lei2020tvr,zhang2021natural} tackle video localization by adapting the concept of extractive question answering and predicting the start and end boundaries of the target moment directly.

The proposal-based framework encompasses proposal-generated and anchor-based methods. Proposal-generated methods~\cite{DBLP:conf/aaai/Xu0PSSS19,xiao2021boundary,xiao2021natural} solve this task through a two-stage \textit{propose-and-rank} pipeline. That is, proposals are first generated, then multimodal matching is employed to predict the most matching proposal for a given query. Anchor-based methods~\cite{NEURIPS2019_6883966f,DBLP:conf/sigir/ZhangLZX19,wang2021structured,DBLP:journals/pami/ZhangPFLL22}, on the other hand, sequentially assign each frame with multiscale temporal anchors and select the anchor with the highest confidence as the result. Among them, 2D-map methods~\cite{wang2021structured,DBLP:journals/pami/ZhangPFLL22} utilize a two-dimensional map to model temporal relations between pre-generated candidate proposals. The map not only enumerates all possible proposals of any length with respect to a time scale/unit, but also maintains their adjacent relations. Consequently, 2D-map based methods have the advantage of leveraging rich contextual information to refine moment representations.

In this paper, the proposed multi-scale 2D temporal map diffusion model makes an alternative exploration to the NLVL task. It introduces the utilization of a generative diffusion model, which facilitates the progressive refinement of 2D score maps. This innovative approach sets our work apart from existing literature in the field.

\subsection{Diffusion Models}

The diffusion model~\cite{sohl2015deep}, originally proposed as a deep latent generative model, has gained significant attention in recent years. Notably, diffusion-based generation has achieved remarkable results for various tasks such as image generation~\cite{dhariwal2021diffusion,rombach2022high}, natural language generation~\cite{gong2022diffuseq,yu2022latent}, and text-to-image synthesis~\cite{gu2022vector,kim2022diffusionclip}. While diffusion models have demonstrated their effectiveness in generative tasks involving images and audio, their potential in understanding tasks has not been thoroughly explored. Some pioneering studies have made initial attempts to apply diffusion models to tasks such as object detection~\cite{chen2022diffusiondet}, image segmentation~\cite{amit2021segdiff,baranchuk2021label}, action segmentation~\cite{liu2023diffusion}, and named entity recognition~\cite{shen2023diffusionner}.  Most relevant to ours, recent works~\cite{li2023momentdiff,zhao2023diffusionvmr} have attempted to adapt diffusion models to  NLVL, using a proposal-free framework and focusing on the start and end locations of target moments as diffusion objectives. Our approach, in contrast, employs a proposal-based framework to adapt diffusion model, and is trained to predict an entire 2D temporal map. Our diffusion objective thus offers a more informative content  representation with continuous values, aligning more closely with the original diffusion formation.
\begin{figure*}
		\centering
		\subfloat[The forward and reverse processes in the proposed diffusion model]{\includegraphics[width=0.48\linewidth]{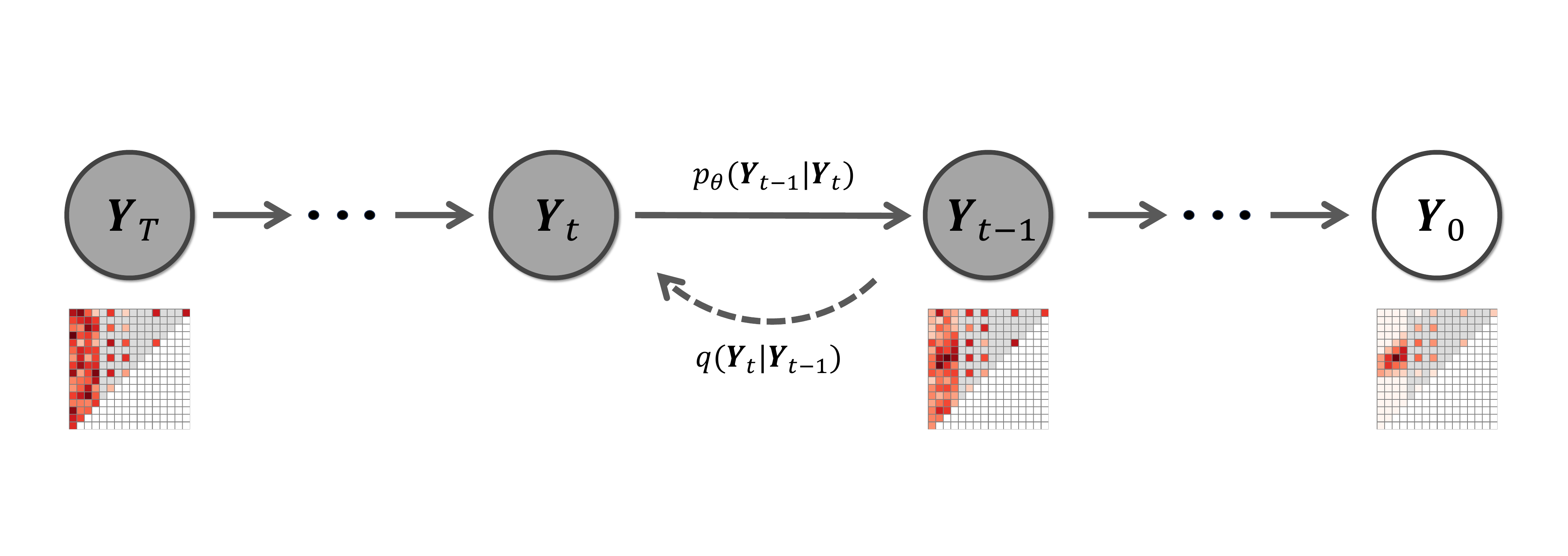}}
		\subfloat[The stylization block for time information interaction]{\includegraphics[width=0.48\linewidth]{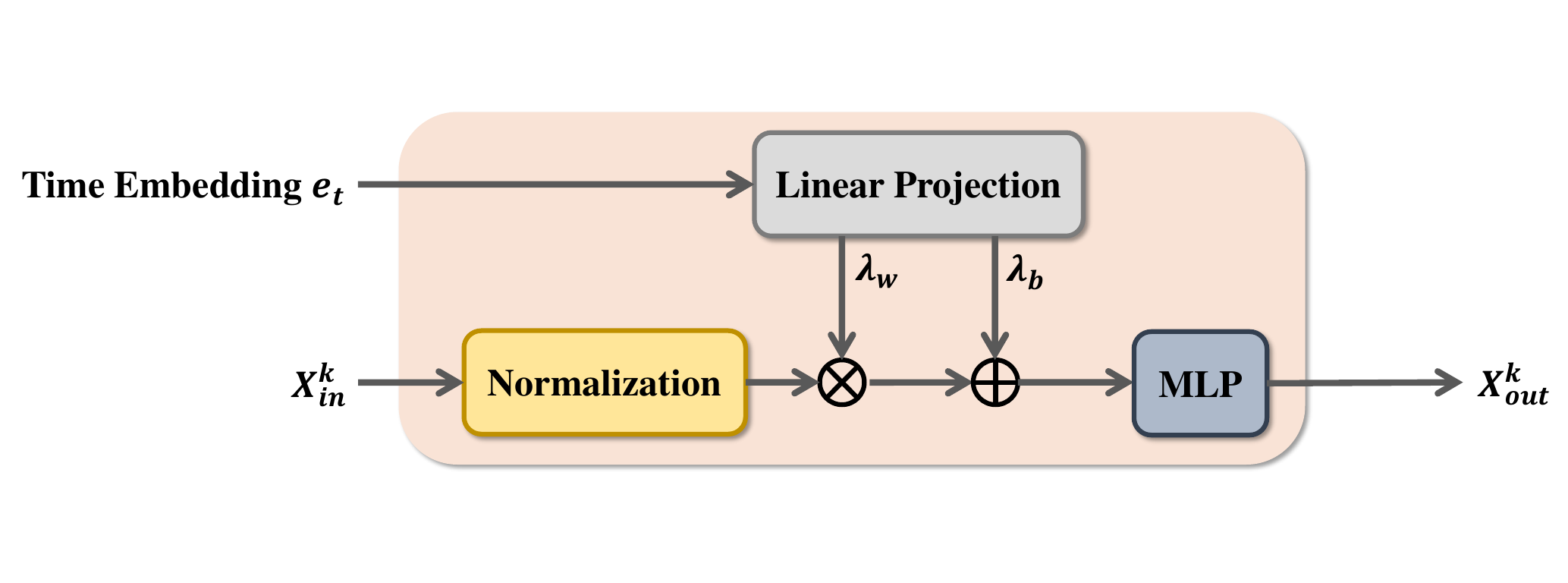}}\\
		\subfloat[The proposed framework: multimodal feature encoder and condition-injected decoder]{\includegraphics[width=0.96\linewidth]{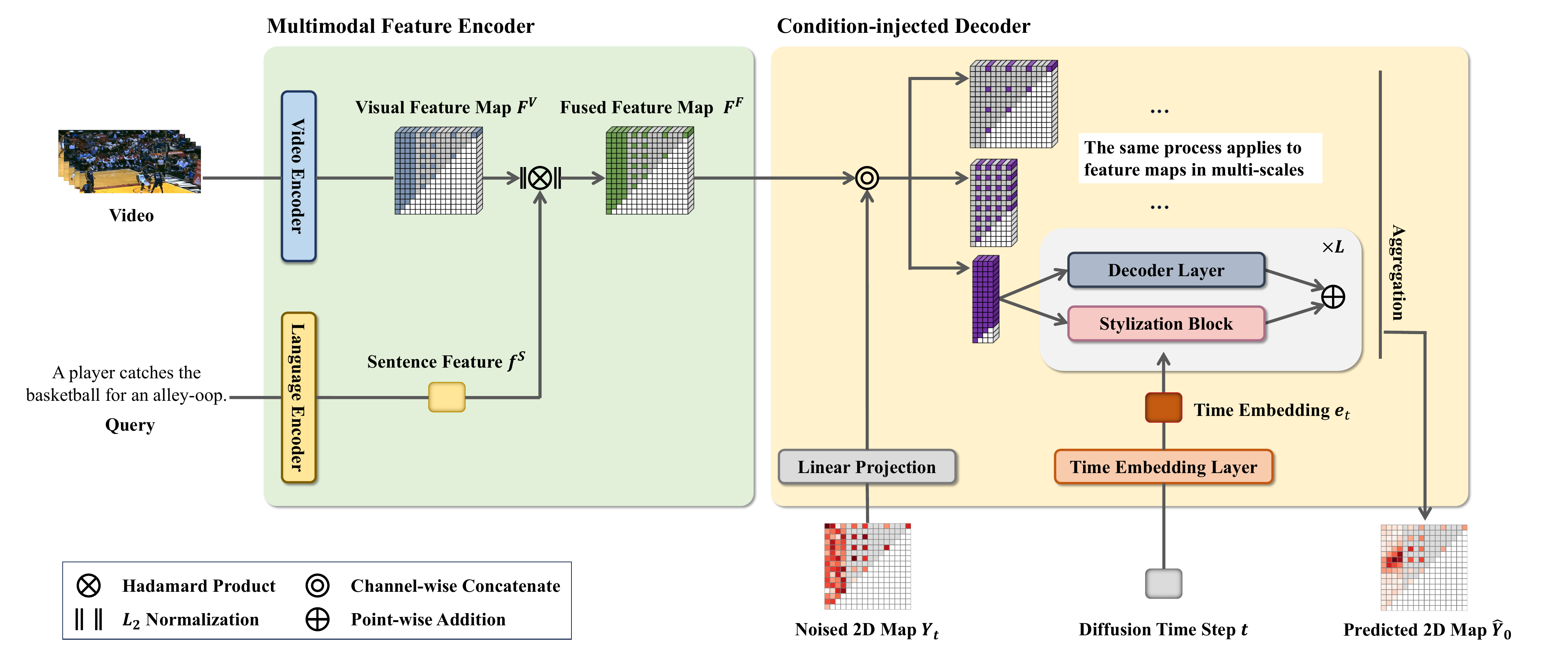}}
	\caption{Overview of our proposed multi-scale 2D temporal map diffusion model. (a) Illustration of the forward and reverse processes in our 2D temporal map-based diffusion model. (b) Design of the stylization block utilized for time information interaction. (c) Framework of the proposed model, incorporating a multimodal feature encoder and a condition-injected decoder.}
    \label{fig:pipeline}
\end{figure*}

\section{Method}
This section begins with the formulation of  2D temporal map in \cref{subsec:problem_definition}. \cref{subsec:2d map diffusion} then describes the diffusion process applied in generating the 2D temporal map. Subsequently, the framework of the proposed diffusion model is outlined in \cref{subsec:framework}. The detailed training and inference procedures are presented in \cref{subsec:training and inference}.

\subsection{2D Temporal Map} \label{subsec:problem_definition}

Let $V$ be an untrimmed video and $S=\{s_i\}_{i=0}^{M-1}$ be a query sentence with $M$ words. The NLVL task is to retrieve a temporal moment $(m_s, m_e)$ on $V$ that semantically corresponds to the query, where $m_s$ and $m_e$ refer to the start and end timestamp of the moment respectively. 

Given the definition, one straightforward formulation is to predict the tuple $(m_s, m_e)$, \textit{i.e.,} the start and end timestamps. However, alternative formulations exist. We follow the 2D-map formulation and utilize a 2D temporal score map, denoted by $\boldsymbol{Y}_{0}$, for moment prediction. To construct this map, we divide the video $V=\{v_i\}_{i=0}^{N-1}$ into $N$ non-overlapping segments each of length $\tau$, using $\tau$ as the time unit. Then there is a correspondence between the segment position and timestamp. Specifically, a candidate moment starting from the timestamp $i\tau$ and with a length of $(j+1)\tau$ can be represented by the segment position $(i, j)$ on a 2D map, illustrated in  \cref{fig:task_intro}. 

The values in the 2D temporal map cells represent the degree of matching between the given sentence and the moment candidates at various $(i, j)$ positions. In our implementation, we utilize the Intersection over Union (IoU) as the matching score. IoU is computed as the ratio of the overlap between a candidate moment and the ground truth moment $(m_s, m_e)$ to their union duration:
\begin{equation}
    \boldsymbol{Y}_{0}^{i,j}=\frac{\left(i\tau,i\tau+(j+1)\tau\right)\cap(m_{s},m_{e})}{(i\tau,i\tau+(j+1)\tau)\cup(m_{s},m_{e})}.
    \label{eq:IoU}
\end{equation}

According to~\cite{DBLP:journals/pami/ZhangPFLL22}, aggregating the results of sparse multi-scale score maps yields higher efficiency and superior performance compared to using a dense single-scale map. Thus, we employ a similar approach in our study. Specifically, we generate $K$ multi-scale maps, where the $k$-th map is extracted from the original single-scale score map at intervals of $2^k$ in both the row and column dimensions. Under this setting, the model is designed to generate $K$ target 2D score maps simultaneously, which will be consolidated into a single-scale map during inference.

\subsection{2D Temporal Map-based Diffusion Process}
\label{subsec:2d map diffusion}

Our approach reinterprets the NLVL task as a conditional generation problem, wherein the input video $V$ and textual query $S$ are considered as conditions, and the objective is to generate the target IoU Map $\boldsymbol{Y}_{0}$. This can be represented as $\hat{\boldsymbol{Y}}_{0}=f_{\theta}(\boldsymbol{Y}_{t},t, V,S)$. The form of $\boldsymbol{Y}_{0}$ can take various variables depending on the specific methodology employed, such as the time step in regression-based methods or the video span in span-based methods. In our case, we opt for utilizing a 2D temporal map as the prediction objective due to its higher informative content, and more importantly, continuous value representation.

We use the Denoising Diffusion Implicit Model (DDIM)~\cite{song2021denoising} for the diffusion process, which consists of two primary components: the \textit{forward} and \textit{reverse} processes.

The forward process is illustrated in \cref{fig:pipeline} (a), which should be viewed from right to left. Given a data distribution $\boldsymbol{Y}_{0} \sim q(\boldsymbol{Y}_{0})$, the \textit{forward} process $q$ progressively incorporate Gaussian noise determined by a predefined variance $\beta_{t} \in (0,1)$ at each time step $t$.\footnote{The time step $t$ here is specific to the diffusion model, not to be confused with the timestamp or time unit specific to the video input.} In this process, it sequentially generates noisy data samples $\boldsymbol{Y}_{1},\boldsymbol{Y}_{2},...,\boldsymbol{Y}_{T}$ as follows:
\begin{equation}
    q(\boldsymbol{Y}_{1: T} | \boldsymbol{Y}_{0}) =\prod_{t=1}^{T} q(\boldsymbol{Y}_{t} | \boldsymbol{Y}_{t-1}),
    \label{eq:foward 0_to_T}
\end{equation}
\begin{equation}
    q(\boldsymbol{Y}_{t} | \boldsymbol{Y}_{t-1})=\mathcal{N}(\boldsymbol{Y}_{t} ; \sqrt{1-\beta_{t}} \boldsymbol{Y}_{t-1}, \beta_{t} \boldsymbol{I}).
    \label{eq:foward t-1_to_t}
\end{equation}

Meanwhile, it is possible to directly sample data $\boldsymbol{Y}_{t}$ at any given time step $t$ without the need for recursively applying \cref{eq:foward t-1_to_t}:
\begin{equation}
    q(\boldsymbol{Y}_{t} | \boldsymbol{Y}_{0})  =\mathcal{N}\left(\boldsymbol{Y}_{t} ; \sqrt{\bar{\alpha}_{t}} \boldsymbol{Y}_{0},(1-\bar{\alpha}_{t}) \boldsymbol{I}\right),
    \label{eq:foward 0_to_t}
\end{equation}
where $\alpha_{t}=1-\beta_{t}$ and $\bar{\alpha}_{t}=\prod_{s=0}^{t} \alpha_{s}$.

The \textit{reverse} process, illustrated in \cref{fig:pipeline} (a) progressing from left to right, involves generating $\boldsymbol{Y}_{0}$ through iterative denoising:
\begin{equation}
\begin{aligned}
   p_{\theta}(\boldsymbol{Y}_{t-1} | \boldsymbol{Y}_{t})  = &\sigma_{t}\epsilon + \sqrt{\bar{\alpha}_{t-1}}f_{\theta}(\boldsymbol{Y}_{t},t, V,S)+ \\
    & \sqrt{1-\bar{\alpha}_{t-1}-\sigma_{t}^{2}}\frac{\boldsymbol{Y}_t-\sqrt{\bar{\alpha}_{t}}f_{\theta}(\boldsymbol{Y}_{t},t, V,S)}{\sqrt{1-\bar{\alpha}_{t}}}.
\end{aligned}
    \label{eq:estimate post t_and_0_to_t-1}
\end{equation}
where $\sigma_{t}$ is controlled by $\beta_{t}$, $\epsilon \in \mathcal{N}(0,\boldsymbol{I})$. Our model $f_{\theta}$ is designed to predict $\boldsymbol{Y}_{0}$ and its prediction is denoted as $\hat{\boldsymbol{Y}}_{0}=f_{\theta}(\boldsymbol{Y}_{t},t, V,S)$. For a more detailed illustration of diffusion theory, we refer readers to~\cite{song2021denoising}.

\subsection{Framework} \label{subsec:framework}
Our proposed model consists of two main components: a \emph{multimodal feature encoder} and a subsequent \emph{condition-injected decoder}, i.e., $f_{\theta}=f_{\theta}^{dec} \circ f_{\theta}^{enc}$. The encoder functions generate multimodal fused feature, denoted as $\bm{F}^{F}$, from both input video segments and query sentences. Then the decoder progressively denoises multi-scale 2D score maps $\boldsymbol{Y}_{t}$, conditioned on $\bm{F}^{F}$. An overview of our framework is presented in \cref{fig:pipeline} (c).

\subsubsection{Multimodal Feature Encoder}

The multimodal feature encoder transforms an input video with $N$ segments $V=\{v_i\}_{i=0}^{N-1}$ and a query sentence containing $M$ words $S=\{s_i\}_{i=0}^{M-1}$ into a multimodal fused feature map $\bm{F}^{F}$, i.e., $\bm{F}^{F}=f_{enc}(V,S)$. We adopt the method in ~\cite{DBLP:journals/pami/ZhangPFLL22} as the main pipeline to generate the fused feature map. Specifically, a visual and text encoder first separately encode the features from the two modalities. Following this, a multimodal feature fusion module is used to generate the fused feature map. The pipeline of the encoder is illustrated in the left part of \cref{fig:pipeline} (c).

Our implementation utilizes a three-layer bidirectional LSTM network as the text encoder.\footnote{The utilization of LSTM aligns with the prevailing approach in NLVL research and contributes to achieving state-of-the-art performance in this task.} It receives a sequence of word embedding vectors, generated by the GloVe  model~\cite{pennington2014glove}, and creates an aggregate sentence feature $\bm{F}^{S} \in \mathbb{R}^{d_{S}}$.

The visual encoder initially processes the video segment sequence $\{v_i\}_{i=0}^{N-1}$ into a feature sequence $\{\bm{f}_{i}^{V}\}_{i=0}^{N-1}$ using a pre-trained 3D CNN and a following fully connected layer.\footnote{Similarly, the utilization of pre-trained 3D CNNs has become a standard practice in NLVL research, showcasing their efficacy and providing a strong foundation for achieving state-of-the-art results in this field.} Subsequently, a stacked convolution module~\cite{zhang2019man} constructs 2D visual feature maps $\bm{F}^V \in \mathbb{R}^{N\times N\times d_{V}}$ from this feature sequence. The dimensions of this map symbolize the start and duration segment indices and the feature dimension. The feature of a moment starting at segment $v_a$ and lasting $(b+1)\tau$ is positioned at $\bm{F}^V[a,b]$.

The multimodal feature fusion module then integrates the 2D visual feature maps $\bm{F}^V$ and the sentence feature $\bm{f}^{S}$. This fusion module first projects these cross-domain features into a shared subspace using fully connected layers. These are then combined via a Hadamard product and $L_2$ normalization. The resultant fused feature map, denoted as $\bm{F}^{F} \in \mathbb{R}^{N\times N\times d_{F}}$, carries a feature dimension of $d_{F}$.

In our implementation, we construct $K$ multi-scale 2D feature maps to match 2D score maps. The $k$-th sparse map is sampled from the fused feature map at intervals of $2^k$ in both the row and column dimensions. We represent the multi-scale feature maps as $\{\bm{F}^{F}_k | \bm{F}^{F}_k \in \mathbb{R}^{N\times 2^k A \times d_{F}}, 0 \leq k \leq K-1\}$, where $A$ represents the number of anchors at each scale. Each map is then processed by different decoders with the same architecture.

\subsubsection{Condition-injected Decoder}

The condition-injected decoder is pivotal in executing the reverse diffusion process. It transforms the noised 2D temporal map, denoted as $\boldsymbol{Y}_t$, into an estimation $\hat{\boldsymbol{Y}}_0$ of the ground truth $\boldsymbol{Y}_0$. This transformation process is conditioned on the multimodal fused feature map $\bm{F}^{F}$ and the time step $t$, mathematically expressed as $\hat{\boldsymbol{Y}}_0=f_{\theta}^{dec}(\boldsymbol{Y}_t,t,\bm{F}^{F})$. As depicted in the right part of \cref{fig:pipeline} (c), the decoder module comprises $K$ decoders, all with identical architecture, operating simultaneously on $K$ multi-scale feature maps. In this section, we detail our design using the pipeline of the $k$-th decoder as an example. \cref{subsec:training and inference} then discuss the aggregation of $K$ predicted 2D score maps into a single-scale map during training and inference respectively.

\begin{itemize}
    \item \textbf{Base Architecture Selection.} As shown in the right part of \cref{fig:pipeline} (c), the decoder contain $L$ stacked blocks and each blcok contains a main function block and a styliztion block. We are here to discuss the architecture selection of the main function block. As depicted in the right part of  \cref{fig:pipeline}(c), the decoder comprises $L$ stacked blocks, with each block consisting of a main function block and a stylization block. This part focuses on the architectural choices for the main function block.
    \begin{itemize}
        \item \textbf{Transformer Model.} One straightforward approach to model design is to leverage successful diffusion architectures from existing generation tasks. In this work, we adopt the transformer architecture proposed in \cite{zhang2022motiondiffuse} as the base diffusion model. To prepare the 2D score map $\boldsymbol{Y}_t^k$ for further processing, it undergoes a transformation through a fully connected layer, resulting in the representation $\boldsymbol{H}_{t}^k$. This transformed representation is fed into the stacked Transformer blocks for subsequent processing and generation. The Transformer block comprises an attention module and a feed-forward network (FFN). The attention module combines self-attention and cross-attention functionalities. Specifically, it utilizes $\boldsymbol{H}_{t}^k$ as the query, while the concatenation of $\boldsymbol{H}_{t}^k$ and $\bm{F}^{F}_k$ serves as the key and value. This design allows the layer to simultaneously enhance the interaction between moments and incorporate the conditioning information into the representation of the score map. Following the processing by the Transformer block, the resulting feature will be utilized for predicting $\hat{\boldsymbol{Y}}_0^k$ using a fully connected layer.
        
        \item \textbf{Convolution Model.} We also investigate a convolution-based decoder in our design, drawing inspiration from the successful implementation of the 2D convolutional architecture described in \cite{DBLP:journals/pami/ZhangPFLL22}. Specifically, we replace the Transformer component with stacked gated convolutional layers \cite{yu2019free}, utilizing large kernel sizes. This architectural choice allows our model to progressively capture more contextual information from neighboring moment candidates while also learning to differentiate between candidate moments. The convolution model is the default architecture in the main experiments.
    \end{itemize}

    \item \textbf{Approaches of Incorporating Conditions.} One crucial consideration in designing a diffusion model is effectively incorporating the conditioning information with the diffusion target, \textit{i.e.,} the 2D temporal score map in our context. 
    \begin{itemize}
        \item \textbf{Cross-attention.} Adopting successful interaction methods from other generation tasks, we utilize cross-attention in a Transformer architecture, as proposed in~\citet{zhang2022motiondiffuse}. This architecture integrates both self-attention and cross-attention functionalities within its attention layer. In this design, the noised 2D temporal map $\boldsymbol{Y}_t^k \in \mathbb{R}^{N \times 2^k A}$ is initially projected to $\boldsymbol{H}_{t}^k \in \mathbb{R}^{N \times 2^k A \times d_{\boldsymbol{H_{t}^k}}}$ using a fully connected layer. Then, $\boldsymbol{H}_{t}^k$ serves as the query, and the concatenation of $\boldsymbol{H}_{t}^k$ and $\bm{F}^{F}_k$ acts as the key and value. This configuration allows the layer to not only enhance interactions between different moments but also integrate conditioning information into the score map representation.

        \item \textbf{Direct Concatenation.} However, the inherent differences between generation and understanding tasks are significant. Generation tasks focus on adhering to given conditions while ensuring diversity in outputs, often without a definitive ground truth for evaluation, thus rely less on conditional information. In contrast, our NLVL task demands the precise generation of predetermined results, necessitating effective leverage of input conditions from both modalities. This requirement suggests that typical interaction mechanisms like plain cross-attention may not be optimal, as supported by findings in \cref{sec:exp_condition}. Consequently, we propose a more direct approach for our NLVL task. This involves processing the concatenation of the condition and projected diffusion target ($[\boldsymbol{H}_{t}^k, \bm{F}^{F}_k]$) together during the denoising process. In the Transformer model, the original attention layer is replaced with a general self-attention layer, using $[\boldsymbol{H}_{t}^k, \bm{F}^{F}_k]$ as the query, key, and value. For the CNN model, this concatenated input is used in the forward pass.
    \end{itemize}
    
    \item \textbf{Time Information Interaction.}
    \begin{itemize}
        \item \textbf{Stylization Block.} Incorporating time step information is essential throughout the denoising process. To facilitate this, we introduce \emph{stylization blocks} within the residual branch, designed to apply a stylization effect on the inputs, as depicted in  \cref{fig:pipeline}(b). The diffusion time step $t$ is initially processed by a time embedding layer, yielding the time embedding $e_t$. The stylization block employs a linear projection to transform $e_t$ into two vectors, $\lambda_{w}$ and $\lambda_{b}$, representing the scale and shift values for the stylization process, respectively. The input to the stylization block, denoted as $X_{in}^k$, is generated from the preceding layer, while $X_{out}^k$ represents its output. Initially, $X_{in}^k$ is normalized using GroupNorm in CNN models, or LayerNorm in Transformer models. Subsequently, it undergoes stylization through $\lambda_{w}$ and $\lambda_{b}$. Finally, a multilayer perceptron (MLP) with a preceding activation function is utilized to generate the output of the stylization block. Thus, the overall process can be mathematically formulated as follows:
        \begin{equation}
            X_{out}^k = \operatorname{MLP}\left(\operatorname{Norm}(X_{in}^k)\cdot \lambda_{w} + \lambda_{b}\right),
            \label{eq:stylization}
        \end{equation}
        where $[\lambda_{w},\lambda_{b}]=\operatorname{Proj}(e_t)$.
        
        \item \textbf{Multi-step Interaction.} To ensure that the time step information has a meaningful influence on the generated outputs, we incorporate a stylization block after every layer. Specifically, in the Transformer block, we place the stylization block after the attention module and the feed-forward network (FFN). As for the CNN model, we position the stylization block at the beginning and after each convolutional layer, respectively. This placement ensures that the time step information is effectively integrated at different stages of the denoising process.
        
    \end{itemize}
    
\end{itemize} 

\subsection{Training and Inference}
\label{subsec:training and inference}

In this section, we elucidate how predictions from different decoders are utilized during the training and inference stages.

Our model $f_{\theta}$ is designed to predict $\boldsymbol{Y}_{0}$. Therefore, in the \emph{training} stage, it is trained to align its prediction $\hat{\boldsymbol{Y}}_{0}=f_{\theta}(\boldsymbol{Y}_{t},t, V,S)$ with the groud truth $\boldsymbol{Y}_{0}$. Since $K$ multi-scale 2D feature maps are used for prediction, we generate corresponding multi-scale noised 2D temporal maps $\{\boldsymbol{Y}_{t}^{k}\}_{k=0}^{K-1}$ via \cref{eq:foward 0_to_t} and the time step $t$ is randomly selected at each training iteration. As for the loss selection, we choose to utilize the original 2D temporal map as the ground truth and employ mean squared error (MSE) for optimization. The training loss is defined as:
\begin{equation}
    \mathcal{L}=\Sigma_{k=0}^{K-1} \mathcal{L}^k,
    \label{eq:loss_total}
\end{equation}
\begin{equation}
    \mathcal{L}^k=\mathbb{E}_{t \sim [1,T],\boldsymbol{Y}_{0} \sim q(\boldsymbol{Y}_{0}),\epsilon \in \mathcal{N}(0,\boldsymbol{I})}[\|f_{\theta}(\boldsymbol{Y}_{t}^{k},t, V,S) - \boldsymbol{Y}_{0}^{k}\|_{2}],
    \label{eq:loss_k}
\end{equation}
where $K$ is the total number of 2D maps. Note that, in MS-2D-TAN~\cite{DBLP:journals/pami/ZhangPFLL22},  a rescaled 2D map and binary cross-entropy (BCE) are used as the prediction objective and training loss, respectively. However, rescaling the 2D map can result in increased sparsity and reduced informativeness. Additionally, the utilization of BCE assumes a discrete prediction objective, which contradicts the continuity requirement in the original DDIM. Hence, we choose differently.  

During the \emph{inference stage}, the model begins with pure noise maps, $\{\boldsymbol{Y}_{T}^{k} \sim \mathcal{N}(0,\boldsymbol{I})\}_{k=0}^{K-1}$, and progressively reduces the noise via \cref{eq:estimate post t_and_0_to_t-1}. Then, all denoised score maps $\{\hat{\boldsymbol{Y}}_{0}^{k}\}_{k=0}^{K-1}$ are recovered to a single-scale map $\hat{\boldsymbol{Y}}_{0}$ based on the moment location at the original single-scale map. For moments that are predicted by more than one score map, we choose the highest score as its final prediction.

\section{Experiment}
\label{sec:exp}

\subsection{Experimental Setting}
\label{ssec:expSettings}

\begin{itemize}
    \item \textbf{Datasets.} Our experiments employ the Charades-STA~\cite{DBLP:conf/iccv/GaoSYN17} and DiDeMo~\cite{anne2017localizing} datasets. The Charades-STA dataset, derived from the original Charades dataset~\cite{DBLP:conf/eccv/SigurdssonVWFLG16}, features $9,848$ videos that depict common indoor activities. This dataset includes $16,128$ moment-language query pairs, with $12,408$ pairs designated for training and $3,720$ for testing. The average lengths of videos and moments are $30.60$ seconds and $8.09$ seconds, respectively. The DiDeMo dataset provides a comprehensive and diverse assortment of videos from Flickr, with a specific focus on event localization through natural language descriptions. The videos, each restricted to maximum 30 seconds, are divided into 5-second segments to facilitate easier annotation. The dataset encompasses $8,395$ training videos, $1,065$ validation, and $1,004$ test videos, resulting in a total of $26,892$ moment-description pairs.

    \item \textbf{Evaluation Metrics.} We follow the evaluation framework of~\cite{DBLP:journals/pami/ZhangPFLL22} for assessing the performance of our model, employing the \emph{$Rank\ n@m$} metric. This metric is derived by determining the percentage of language queries where at least one \textit{accurate moment retrieval} appears within the top-$n$ retrieved moments. A moment retrieval is considered \textit{accurate} if its Intersection over Union (IoU) with the ground truth moment exceeds the threshold $m$. The specific combinations for $n$ and $m$ may differ between datasets.
    For both Charades-STA and DiDeMo datasets, we report results with $n\in \{1,5\}$ and $m\in \{0.5,0.7\}$.
\end{itemize}

\begin{table}[t]
\centering
\resizebox{1.0\linewidth}{!}{
\begin{tabular}{c|c|cccc}
\toprule
\multirow{2}{*}{\textbf{Method}} & \multirow{2}{*}{\textbf{Feature}} & \multicolumn{2}{c}{$\bm{Rank1@}$} & \multicolumn{2}{c}{$\bm{Rank5@}$} \\
\cmidrule(r){3-4} \cmidrule(r){5-6}
~ & ~ & $\bm{0.5}$ & $\bm{0.7}$ & $\bm{0.5}$ & $\bm{0.7}$ \\
\midrule 
LGI \cite{mun2020local} & I3D & 59.5 & 35.5 & - & -  \\
VSLNet \cite{DBLP:conf/acl/ZhangSJZ20} & I3D & 54.2 & 35.2 & - & -  \\
IVG-DCL \cite{nan2021interventional} & I3D & 50.2 & 32.9 & - & - \\
ACRM \cite{tang2021frame} & I3D & 57.5 & 38.3 & - & -  \\
GTR-H \cite{DBLP:conf/emnlp/CaoCSZZ21} & RAW & 62.6 & 39.7 & 91.6 & 62.0 \\
\midrule 
CTRL \cite{DBLP:conf/iccv/GaoSYN17} & C3D & 23.6 & 8.9 & 58.9 & 29.6 \\
QSPN \cite{DBLP:conf/aaai/Xu0PSSS19} & C3D & 35.6 & 15.8 & 79.4 & 45.5 \\
2D-TAN \cite{zhang2020learning} & VGG & 39.8 & 23.3 & 79.3 & 51.2 \\
BPNet \cite{xiao2021boundary} & I3D & 50.8 & 31.6 & - & - \\ 
SCDM \cite{NEURIPS2019_6883966f} & I3D & 54.4 & 33.4 & 74.4 & 58.1 \\
MS-2D-TAN \cite{DBLP:journals/pami/ZhangPFLL22} & I3D & 60.1 & 37.4 & 89.1 & 59.2  \\
\midrule 
\textbf{Ours} & I3D & 60.3 & 40.8 & 79.7 & 63.0  \\
\bottomrule
\end{tabular}
}
\caption{
Performance  on Charades-STA. Results of other models are obtained from their original papers.
}
\label{tab:charades}
\end{table}

\begin{table}[t]
\centering
\resizebox{1.0\linewidth}{!}{
\begin{tabular}{c|c|cccc}
\toprule
\multirow{2}{*}{\textbf{Method}} & \multirow{2}{*}{\textbf{Feature}} & \multicolumn{2}{c}{$\bm{Rank1@}$} & \multicolumn{2}{c}{$\bm{Rank5@}$} \\
\cmidrule(r){3-4} \cmidrule(r){5-6}
~ & ~ & $\bm{0.5}$ & $\bm{0.7}$ & $\bm{0.5}$ & $\bm{0.7}$ \\
\midrule 
ACRN \cite{liu2018attentive} & VGG & 27.4 & 16.7 & 69.4 & 29.5 \\
CSMGAN \cite{liu2020jointly} & C3D & 29:4 & 19.2 & 70.8 & 41.6 \\
VLG-Net \cite{soldan2021vlg} & VGG & 33.4 & 25.6 & 88.7 & 71.7 \\
MS-2D-TAN \cite{DBLP:journals/pami/ZhangPFLL22} & VGG & 26.1 & 23.0 & 76.1 & 63.6  \\
\midrule 
\textbf{Ours} & VGG & 31.5 & 25.3 & 72.1 & 65.4  \\
\bottomrule
\end{tabular}
}
\caption{
Performance on DiDeMo. Results of other models are obtained from their original papers, except MS-2D-TAN, which are obtained in our experiments.
}
\label{tab:didemo}
\end{table}

\subsection{Performance Comparison}

We compare our CNN-based diffusion model and other state-of-the-art models. The evaluation results on the Charades-STA dataset are outlined in \cref{tab:charades}. Notably, our model demonstrates exceptional performance, surpassing other models in both top-1 and top-5 prediction measures when $m=0.7$. This enhancement is particularly remarkable contrasting the outcomes with MS-2D-TAN. 

The core of our model is an iterative diffusion process which starts from a pure noise state and progressively denoises it, resulting in a 2D temporal map of higher accuracy than those generated by the single inference processes typical in discriminative models. This improvement underscores the potency of diffusion models in tasks requiring deep understanding, especially when they are adeptly designed to exploit multi-modal conditions. The success of our model in these respects indicates that our diffusion model architecture is not only well-suited but also excels in the task of NLVL.

In \cref{tab:didemo}, we observe that the original MS-2D-TAN model performs poorly as a strong baseline model on the DiDeMo dataset. However, using the same convolution architecture, our diffusion model outperforms it on 3 out of 4 measures. This suggests that the diffusion model has the potential to consistently enhance performance across different datasets.

\begin{table}[t]
\centering
\resizebox{1.0\linewidth}{!}{
\begin{threeparttable}
\begin{tabular}{c|c|cccc}
\toprule
\multirow{2}{*}{\textbf{Architecure}} & \multirow{2}{*}{\textbf{Approach}} & \multicolumn{2}{c}{$\bm{Rank1@}$} & \multicolumn{2}{c}{$\bm{Rank5@}$} \\
\cmidrule(r){3-4} \cmidrule(r){5-6}
~ & ~ & $\bm{0.5}$ & $\bm{0.7}$ & $\bm{0.5}$ & $\bm{0.7}$ \\
\midrule
Transformer & x-attn + ss & 37.0 & 20.2 & 59.4 & 38.4   \\
Transformer & x-attn & 40.8 & 22.3 & 64.0 & 43.0   \\
Transformer & concat &  53.2 & 32.1 & 72.0 & 50.3   \\
\midrule 
CNN & mul & 57.9 & 37.0 & 79.1 & 57.7  \\
CNN & concat & 60.3 & 40.8 & 79.7 & 63.0  \\
\bottomrule
\end{tabular}
\begin{tablenotes}
\item \textbf{x-attn}: cross-attention; \textbf{ss}: single-scale; \textbf{concat}: concatenation; \textbf{mul}: multiplication.
\end{tablenotes}
\end{threeparttable}
}
\caption{Ablation study on condition incorporation methods on the Charades-STA dataset. Results indicate that the concatenation approach in condition interaction yields best performance.
}
\label{tab:ablation_condition}
\end{table}

\begin{table}[t]
\centering
\resizebox{1.0\linewidth}{!}{
\begin{tabular}{c|c|cccc}
\toprule
\multirow{2}{*}{\textbf{Loss Type}} & \multirow{2}{*}{\textbf{2D Map Type}} & \multicolumn{2}{c}{$\bm{Rank1@}$} & \multicolumn{2}{c}{$\bm{Rank5@}$} \\
\cmidrule(r){3-4} \cmidrule(r){5-6}
~ & ~ & $\bm{0.5}$ & $\bm{0.7}$ & $\bm{0.5}$ & $\bm{0.7}$ \\
\midrule 
BCE & rescaled map & 56.6 & 39.5 & 84.2 & 61.3   \\
BCE & full map &  58.6 & 38.3 & 79.1 & 58.6   \\
MSE & full map & 60.3 & 40.8 & 79.7 & 63.0   \\
\bottomrule
\end{tabular}
}
\caption{
Ablation study on optimization objective design on the Charades-STA dataset. Results show that integrating mean squared error (MSE) loss with full 2D score map improves the generation objective and optimizes the diffusion process.
}
\label{tab:ablation_loss}
\end{table}

\begin{table}[t]
\centering
\resizebox{1.0\linewidth}{!}{
\begin{tabular}{cc|cccc}
\toprule
\multirow{2}{*}{\textbf{Fusion Layer}} & \multirow{2}{*}{\textbf{Conv Layer}} & \multicolumn{2}{c}{$\bm{Rank1@}$} & \multicolumn{2}{c}{$\bm{Rank5@}$} \\
\cmidrule(r){3-4} \cmidrule(r){5-6}
~ & ~ & $\bm{0.5}$ & $\bm{0.7}$ & $\bm{0.5}$ & $\bm{0.7}$ \\
\midrule 
- stylization & - stylization  & 59.1 & 37.4 & 81.5 & 57.7   \\
+ stylization & - stylization & 60.2 & 39.0 & 80.4 & 59.3  \\
- stylization & + stylization & 59.7 & 38.4 & 79.9 & 58.6  \\
+ stylization & + stylization & 60.3 & 40.8 & 79.7 & 63.0  \\
\bottomrule
\end{tabular}
}
\caption{
Ablation Study on Time Information Interaction. Symbols + and - indicate with and without stylization blocks in the corresponding layer. Findings indicate that stylization blocks at any position improve performance, with the best results achieved when all blocks are active, underscoring the value of multi-step temporal interaction for denoising 2D temporal maps.
}
\label{tab:ablation_sty_blk}
\end{table}

\begin{figure*}[t!]
\centering
\subfloat[Sample of correct prediction]{\includegraphics[width=0.48\linewidth]{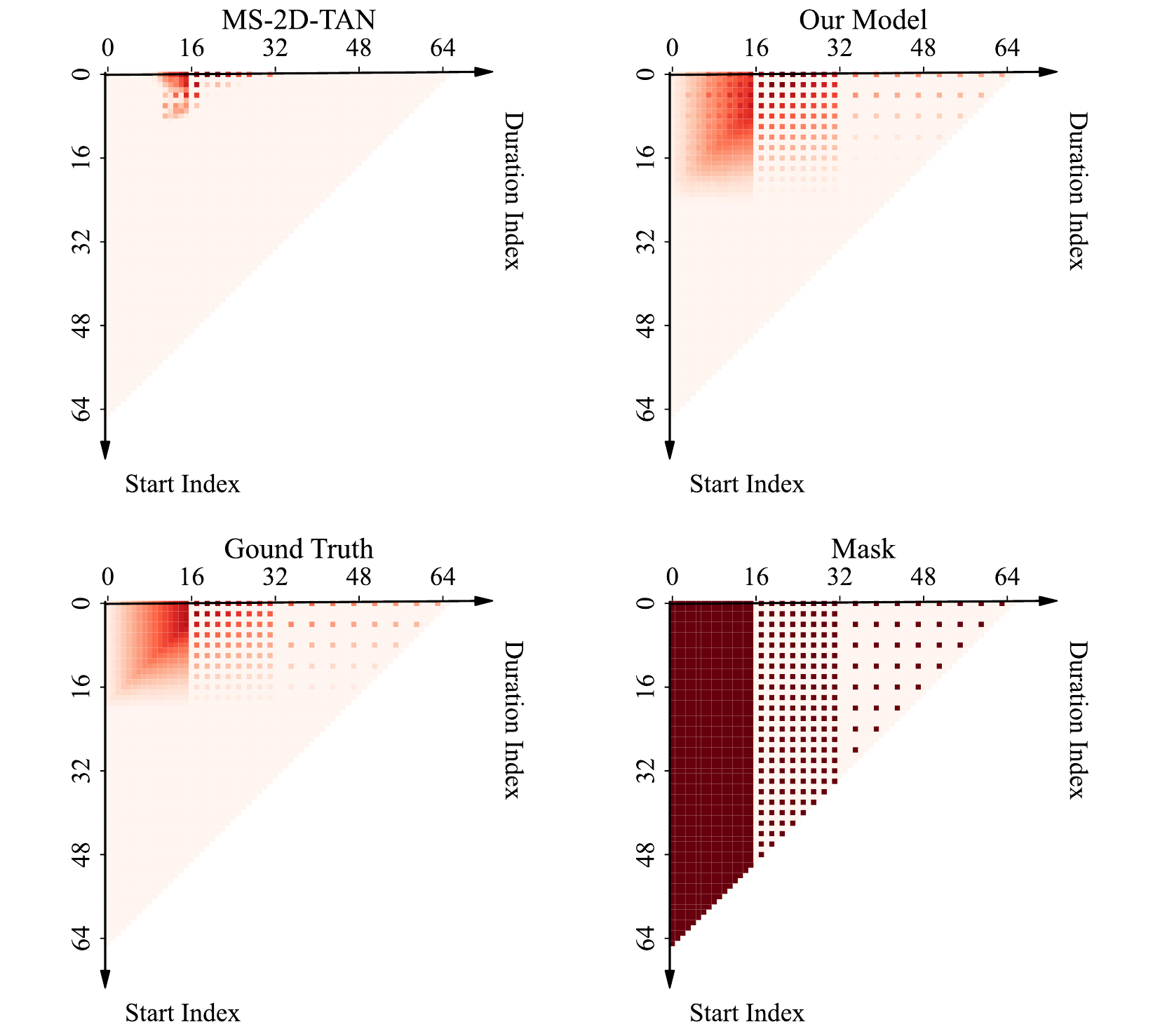}}
\hfill
\subfloat[Sample of wrong prediction]{\includegraphics[width=0.48\linewidth]{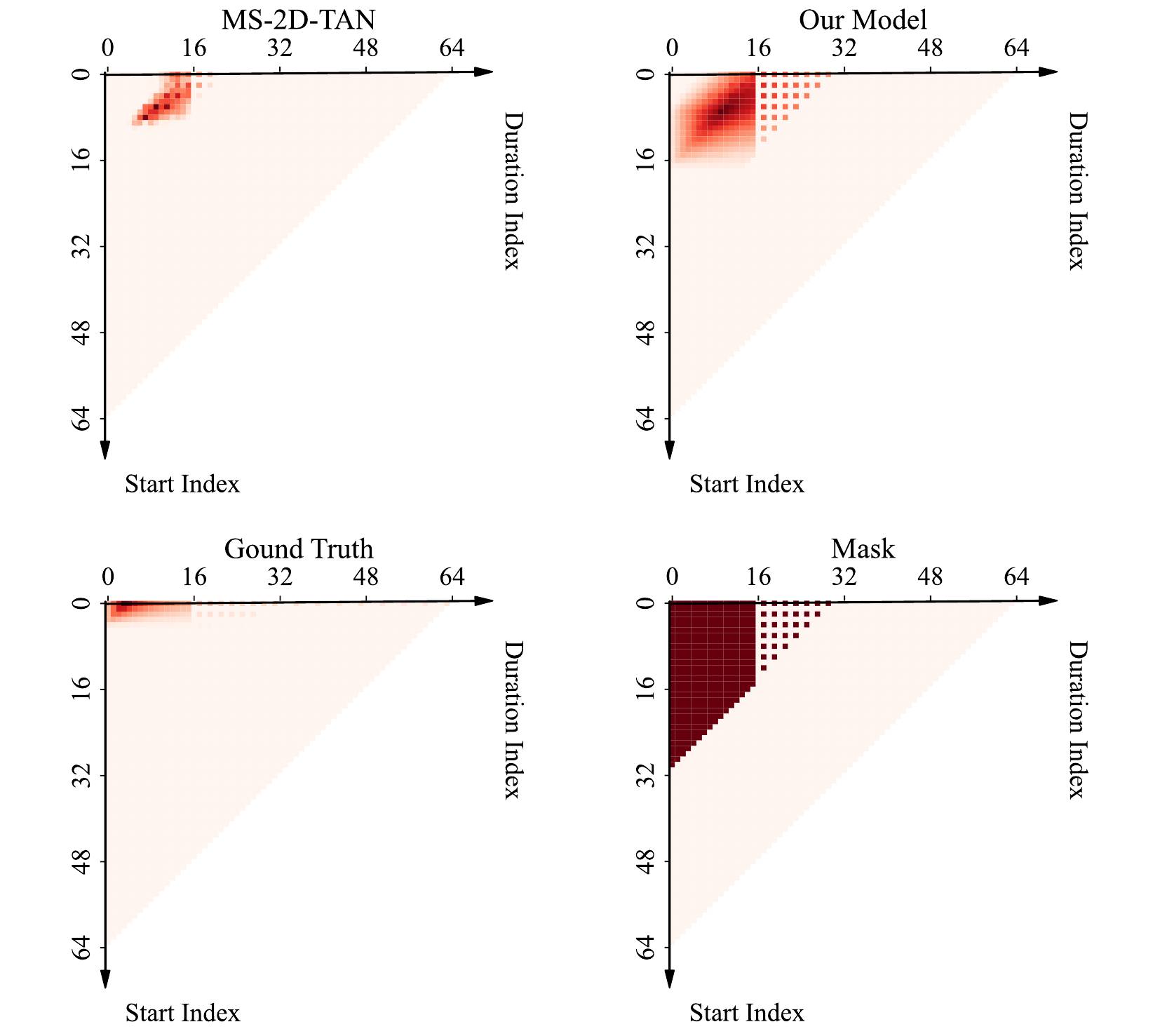}}

\caption{Visualizations of predicted 2D maps for two samples (from Charades-STA), generated by the MS-2D-TAN model and our diffusion model. The diffusion model consistently produces 2D maps with a recognizable pattern, despite  occasional incorrect predictions.}
\label{fig: charades_2d_vis}
\end{figure*}

\subsection{Ablation Study}
\label{ssec:expAblation}

\subsubsection{Approaches of Incorporating Conditions} \label{sec:exp_condition}

As part of our initial investigation, we evaluated various strategies for introducing conditional information into the denoising process. For Transformer-based models, we experimented with integrating the conditioning information into the diffusion target via both \emph{cross-attention} mechanisms and \emph{concatenation}. With cross-attention, we not only processed multi-scale 2D maps separately but also combined them into a single sequence, referred to as \emph{single-scale}. In the context of CNN models, we examined several alternatives, including different \emph{multiplicative} interactions between the score map and the feature map for enhanced denoising. Among many experiments conducted, we only list the best results obtained on the Charades-STA dataset here.  \cref{tab:ablation_condition} reveals that the concatenation method outperforms others with both Transformer and CNN frameworks. This underscores the indispensable role of conditional information in understanding-oriented tasks for making precise predictions. Consequently, our findings confirm the effectiveness of the concatenation approach in condition interaction for optimal performance in NLVL tasks.

In addition, also observe in \cref{tab:ablation_condition} that the CNN-based diffusion model outperforms its Transformer-based counterpart. This superior performance is primarily attributed to the inductive bias of the convolutional model. It facilitates the progressive capture of contextual information, starting from neighboring moments and extending across all candidates. In contrast, the Transformer model processes all moments simultaneously from the input layer, which hinders the model's ability to effectively differentiate between adjacent moments.

\subsubsection{Design of Optimization Objective}

To validate the soundness of our chosen optimization objective, we executed a series of experiments examining various combinations of loss functions and 2D map configurations. The findings, reported in \cref{tab:ablation_loss}, indicate that our proposed approach surpasses competing methods on most measures. This substantiates our initial premise that the application of mean squared error (MSE) loss in conjunction with a full 2D score map constitutes a more informative generative objective. Such an objective is demonstrably more conducive to the diffusion process, aligning with the generative nature of the task at hand.

\subsubsection{Time Information Interaction}

We additionally evaluated the impact of diffusion time step interactions on our model's performance. In the control experiments, we systematically removed stylization blocks from various positions within the architecture, as indicated in \cref{tab:ablation_sty_blk}. A comparative analysis between the baseline (first row) and subsequent configurations (following rows) reveals that the inclusion of stylization blocks in both layers contributes to improved performance. Notably, the model exhibits peak performance when all stylization blocks are operational, signifying the advantage of multi-step temporal information interaction. These findings corroborate the efficacy of stylization blocks in infusing time step information into the denoising process of 2D temporal maps.

\subsection{Qualitative Result}

In \cref{fig: charades_2d_vis}, we present visualizations of the predicted 2D maps generated by both the MS-2D-TAN model and our diffusion model alongside the ground truth. Upon comparing these visualizations, it becomes evident that the diffusion model produces 2D maps with a consistent pattern, although the predictions may not be entirely correct. Specifically, the diffusion model tends to generate a plausible distribution (values decrease when they get far from the predicted center), even when the center of the heat map is incorrectly located. In contrast, the quality of the 2D maps generated by the MS-2D-TAN model is subpar, even when the top-1 prediction is accurate.
\section{Conclusion}
In this study, we redefine the NLVL task as a conditional diffusion generation challenge, focusing on creating a global 2D temporal map to better capture temporal dynamics. We identify key differences between generative and understanding tasks, and particularly in the context of NLVL. We then develop a specialized diffusion decoder tailored for NLVL, incorporating conditions and diffusion time steps. The main  modifications include altering the base architecture, refining the integration of conditions, enhancing the interaction with time information, and designing optimization objective. These changes resulted in improved feature modeling and a more suitable diffusion process for the NLVL task. Experiments confirmed the effectiveness of our design, highlighting the unique characteristics of our diffusion model. We believe this work offers a fresh perspective on utilizing diffusion models in multimodal understanding tasks and could serve as a guide for future innovations in this field.
{

    \small
    \bibliographystyle{ieeenat_fullname}
    \bibliography{main}

\begin{thebibliography}{44}
\providecommand{\natexlab}[1]{#1}
\providecommand{\url}[1]{\texttt{#1}}
\expandafter\ifx\csname urlstyle\endcsname\relax
  \providecommand{\doi}[1]{doi: #1}\else
  \providecommand{\doi}{doi: \begingroup \urlstyle{rm}\Url}\fi

\bibitem[Amit et~al.(2021)Amit, Shaharbany, Nachmani, and Wolf]{amit2021segdiff}
Tomer Amit, Tal Shaharbany, Eliya Nachmani, and Lior Wolf.
\newblock Segdiff: Image segmentation with diffusion probabilistic models.
\newblock \emph{arXiv preprint arXiv:2112.00390}, 2021.

\bibitem[Anne~Hendricks et~al.(2017)Anne~Hendricks, Wang, Shechtman, Sivic, Darrell, and Russell]{anne2017localizing}
Lisa Anne~Hendricks, Oliver Wang, Eli Shechtman, Josef Sivic, Trevor Darrell, and Bryan Russell.
\newblock Localizing moments in video with natural language.
\newblock In \emph{ICCV}, pages 5803--5812, 2017.

\bibitem[Baranchuk et~al.(2021)Baranchuk, Rubachev, Voynov, Khrulkov, and Babenko]{baranchuk2021label}
Dmitry Baranchuk, Ivan Rubachev, Andrey Voynov, Valentin Khrulkov, and Artem Babenko.
\newblock Label-efficient semantic segmentation with diffusion models.
\newblock \emph{arXiv preprint arXiv:2112.03126}, 2021.

\bibitem[Cao et~al.(2021)Cao, Chen, Shou, Zhang, and Zou]{DBLP:conf/emnlp/CaoCSZZ21}
Meng Cao, Long Chen, Mike~Zheng Shou, Can Zhang, and Yuexian Zou.
\newblock On pursuit of designing multi-modal transformer for video grounding.
\newblock In \emph{EMNLP}, pages 9810--9823. Association for Computational Linguistics, 2021.

\bibitem[Chen et~al.(2020)Chen, Lu, Tang, Xiao, Zhang, Tan, and Li]{DBLP:conf/aaai/ChenLTXZTL20}
Long Chen, Chujie Lu, Siliang Tang, Jun Xiao, Dong Zhang, Chilie Tan, and Xiaolin Li.
\newblock Rethinking the bottom-up framework for query-based video localization.
\newblock In \emph{AAAI}, pages 10551--10558. {AAAI} Press, 2020.

\bibitem[Chen et~al.(2022)Chen, Sun, Song, and Luo]{chen2022diffusiondet}
Shoufa Chen, Peize Sun, Yibing Song, and Ping Luo.
\newblock Diffusiondet: Diffusion model for object detection.
\newblock \emph{arXiv preprint arXiv:2211.09788}, 2022.

\bibitem[Dhariwal and Nichol(2021)]{dhariwal2021diffusion}
Prafulla Dhariwal and Alexander Nichol.
\newblock Diffusion models beat gans on image synthesis.
\newblock \emph{NeurIPS}, 34:\penalty0 8780--8794, 2021.

\bibitem[Gao et~al.(2017)Gao, Sun, Yang, and Nevatia]{DBLP:conf/iccv/GaoSYN17}
Jiyang Gao, Chen Sun, Zhenheng Yang, and Ram Nevatia.
\newblock {TALL:} temporal activity localization via language query.
\newblock In \emph{ICCV}, pages 5277--5285. {IEEE} Computer Society, 2017.

\bibitem[Ghosh et~al.(2019)Ghosh, Agarwal, Parekh, and Hauptmann]{ghosh2019excl}
Soham Ghosh, Anuva Agarwal, Zarana Parekh, and Alexander Hauptmann.
\newblock Excl: Extractive clip localization using natural language descriptions.
\newblock \emph{arXiv preprint arXiv:1904.02755}, 2019.

\bibitem[Gong et~al.(2022)Gong, Li, Feng, Wu, and Kong]{gong2022diffuseq}
Shansan Gong, Mukai Li, Jiangtao Feng, Zhiyong Wu, and LingPeng Kong.
\newblock Diffuseq: Sequence to sequence text generation with diffusion models.
\newblock \emph{arXiv preprint arXiv:2210.08933}, 2022.

\bibitem[Gu et~al.(2022)Gu, Chen, Bao, Wen, Zhang, Chen, Yuan, and Guo]{gu2022vector}
Shuyang Gu, Dong Chen, Jianmin Bao, Fang Wen, Bo Zhang, Dongdong Chen, Lu Yuan, and Baining Guo.
\newblock Vector quantized diffusion model for text-to-image synthesis.
\newblock In \emph{CVPR}, pages 10696--10706, 2022.

\bibitem[Kim et~al.(2022)Kim, Kwon, and Ye]{kim2022diffusionclip}
Gwanghyun Kim, Taesung Kwon, and Jong~Chul Ye.
\newblock Diffusionclip: Text-guided diffusion models for robust image manipulation.
\newblock In \emph{CVPR}, pages 2426--2435, 2022.

\bibitem[Lei et~al.(2020)Lei, Yu, Berg, and Bansal]{lei2020tvr}
Jie Lei, Licheng Yu, Tamara~L Berg, and Mohit Bansal.
\newblock Tvr: A large-scale dataset for video-subtitle moment retrieval.
\newblock In \emph{ECCV}, pages 447--463. Springer, 2020.

\bibitem[Li et~al.(2023)Li, Xie, Xie, Zhao, Zhang, Zheng, Zhao, and Zhang]{li2023momentdiff}
Pandeng Li, Chen-Wei Xie, Hongtao Xie, Liming Zhao, Lei Zhang, Yun Zheng, Deli Zhao, and Yongdong Zhang.
\newblock Momentdiff: Generative video moment retrieval from random to real.
\newblock In \emph{NeurIPS}, 2023.

\bibitem[Liu et~al.(2020)Liu, Qu, Liu, Dong, Zhou, and Xu]{liu2020jointly}
Daizong Liu, Xiaoye Qu, Xiao-Yang Liu, Jianfeng Dong, Pan Zhou, and Zichuan Xu.
\newblock Jointly cross-and self-modal graph attention network for query-based moment localization.
\newblock In \emph{ACM MM}, pages 4070--4078, 2020.

\bibitem[Liu et~al.(2023)Liu, Li, Dinh, Jiang, Shah, and Xu]{liu2023diffusion}
Daochang Liu, Qiyue Li, AnhDung Dinh, Tingting Jiang, Mubarak Shah, and Chang Xu.
\newblock Diffusion action segmentation.
\newblock \emph{arXiv preprint arXiv:2303.17959}, 2023.

\bibitem[Liu et~al.(2018)Liu, Wang, Nie, He, Chen, and Chua]{liu2018attentive}
Meng Liu, Xiang Wang, Liqiang Nie, Xiangnan He, Baoquan Chen, and Tat-Seng Chua.
\newblock Attentive moment retrieval in videos.
\newblock In \emph{SIGIR}, pages 15--24, 2018.

\bibitem[Lu et~al.(2019)Lu, Chen, Tan, Li, and Xiao]{lu2019debug}
Chujie Lu, Long Chen, Chilie Tan, Xiaolin Li, and Jun Xiao.
\newblock Debug: A dense bottom-up grounding approach for natural language video localization.
\newblock In \emph{EMNLP}, pages 5144--5153, 2019.

\bibitem[Mun et~al.(2020)Mun, Cho, and Han]{mun2020local}
Jonghwan Mun, Minsu Cho, and Bohyung Han.
\newblock Local-global video-text interactions for temporal grounding.
\newblock In \emph{CVPR}, pages 10810--10819, 2020.

\bibitem[Nan et~al.(2021)Nan, Qiao, Xiao, Liu, Leng, Zhang, and Lu]{nan2021interventional}
Guoshun Nan, Rui Qiao, Yao Xiao, Jun Liu, Sicong Leng, Hao Zhang, and Wei Lu.
\newblock Interventional video grounding with dual contrastive learning.
\newblock In \emph{CVPR}, pages 2765--2775, 2021.

\bibitem[Pennington et~al.(2014)Pennington, Socher, and Manning]{pennington2014glove}
Jeffrey Pennington, Richard Socher, and Christopher~D Manning.
\newblock Glove: Global vectors for word representation.
\newblock In \emph{EMNLP}, pages 1532--1543, 2014.

\bibitem[Rombach et~al.(2022)Rombach, Blattmann, Lorenz, Esser, and Ommer]{rombach2022high}
Robin Rombach, Andreas Blattmann, Dominik Lorenz, Patrick Esser, and Bj{\"o}rn Ommer.
\newblock High-resolution image synthesis with latent diffusion models.
\newblock In \emph{CVPR}, pages 10684--10695, 2022.

\bibitem[Shen et~al.(2023)Shen, Song, Tan, Li, Lu, and Zhuang]{shen2023diffusionner}
Yongliang Shen, Kaitao Song, Xu Tan, Dongsheng Li, Weiming Lu, and Yueting Zhuang.
\newblock Diffusionner: Boundary diffusion for named entity recognition.
\newblock \emph{arXiv preprint arXiv:2305.13298}, 2023.

\bibitem[Sigurdsson et~al.(2016)Sigurdsson, Varol, Wang, Farhadi, Laptev, and Gupta]{DBLP:conf/eccv/SigurdssonVWFLG16}
Gunnar~A. Sigurdsson, G{\"{u}}l Varol, Xiaolong Wang, Ali Farhadi, Ivan Laptev, and Abhinav Gupta.
\newblock Hollywood in homes: Crowdsourcing data collection for activity understanding.
\newblock In \emph{ECCV}, pages 510--526. Springer, 2016.

\bibitem[Sohl-Dickstein et~al.(2015)Sohl-Dickstein, Weiss, Maheswaranathan, and Ganguli]{sohl2015deep}
Jascha Sohl-Dickstein, Eric Weiss, Niru Maheswaranathan, and Surya Ganguli.
\newblock Deep unsupervised learning using nonequilibrium thermodynamics.
\newblock In \emph{ICML}, pages 2256--2265. PMLR, 2015.

\bibitem[Soldan et~al.(2021)Soldan, Xu, Qu, Tegner, and Ghanem]{soldan2021vlg}
Mattia Soldan, Mengmeng Xu, Sisi Qu, Jesper Tegner, and Bernard Ghanem.
\newblock Vlg-net: Video-language graph matching network for video grounding.
\newblock In \emph{ICCV}, pages 3224--3234, 2021.

\bibitem[Song et~al.(2021)Song, Meng, and Ermon]{song2021denoising}
Jiaming Song, Chenlin Meng, and Stefano Ermon.
\newblock Denoising diffusion implicit models.
\newblock In \emph{ICLR}, 2021.

\bibitem[Tang et~al.(2021)Tang, Zhu, Liu, Gao, and Cheng]{tang2021frame}
Haoyu Tang, Jihua Zhu, Meng Liu, Zan Gao, and Zhiyong Cheng.
\newblock Frame-wise cross-modal matching for video moment retrieval.
\newblock \emph{IEEE TMM}, 24:\penalty0 1338--1349, 2021.

\bibitem[Wang et~al.(2021)Wang, Zha, Li, Liu, and Luo]{wang2021structured}
Hao Wang, Zheng-Jun Zha, Liang Li, Dong Liu, and Jiebo Luo.
\newblock Structured multi-level interaction network for video moment localization via language query.
\newblock In \emph{CVPR}, pages 7026--7035, 2021.

\bibitem[Xiao et~al.(2021{\natexlab{a}})Xiao, Chen, Shao, Zhuang, and Xiao]{xiao2021natural}
Shaoning Xiao, Long Chen, Jian Shao, Yueting Zhuang, and Jun Xiao.
\newblock Natural language video localization with learnable moment proposals.
\newblock \emph{arXiv preprint arXiv:2109.10678}, 2021{\natexlab{a}}.

\bibitem[Xiao et~al.(2021{\natexlab{b}})Xiao, Chen, Zhang, Ji, Shao, Ye, and Xiao]{xiao2021boundary}
Shaoning Xiao, Long Chen, Songyang Zhang, Wei Ji, Jian Shao, Lu Ye, and Jun Xiao.
\newblock Boundary proposal network for two-stage natural language video localization.
\newblock In \emph{AAAI}, pages 2986--2994, 2021{\natexlab{b}}.

\bibitem[Xu et~al.(2019)Xu, He, Plummer, Sigal, Sclaroff, and Saenko]{DBLP:conf/aaai/Xu0PSSS19}
Huijuan Xu, Kun He, Bryan~A. Plummer, Leonid Sigal, Stan Sclaroff, and Kate Saenko.
\newblock Multilevel language and vision integration for text-to-clip retrieval.
\newblock In \emph{AAAI}, pages 9062--9069. {AAAI} Press, 2019.

\bibitem[Yu et~al.(2019)Yu, Lin, Yang, Shen, Lu, and Huang]{yu2019free}
Jiahui Yu, Zhe Lin, Jimei Yang, Xiaohui Shen, Xin Lu, and Thomas~S Huang.
\newblock Free-form image inpainting with gated convolution.
\newblock In \emph{CVPR}, pages 4471--4480, 2019.

\bibitem[Yu et~al.(2022)Yu, Xie, Ma, Jia, Pang, Gao, Zhu, Zhu, and Wu]{yu2022latent}
Peiyu Yu, Sirui Xie, Xiaojian Ma, Baoxiong Jia, Bo Pang, Ruigi Gao, Yixin Zhu, Song-Chun Zhu, and Ying~Nian Wu.
\newblock Latent diffusion energy-based model for interpretable text modeling.
\newblock \emph{arXiv preprint arXiv:2206.05895}, 2022.

\bibitem[Yuan et~al.(2019{\natexlab{a}})Yuan, Ma, Wang, Liu, and Zhu]{NEURIPS2019_6883966f}
Yitian Yuan, Lin Ma, Jingwen Wang, Wei Liu, and Wenwu Zhu.
\newblock Semantic conditioned dynamic modulation for temporal sentence grounding in videos.
\newblock In \emph{NeurIPS}. Curran Associates, Inc., 2019{\natexlab{a}}.

\bibitem[Yuan et~al.(2019{\natexlab{b}})Yuan, Mei, and Zhu]{yuan2019find}
Yitian Yuan, Tao Mei, and Wenwu Zhu.
\newblock To find where you talk: Temporal sentence localization in video with attention based location regression.
\newblock In \emph{AAAI}, pages 9159--9166, 2019{\natexlab{b}}.

\bibitem[Zhang et~al.(2019{\natexlab{a}})Zhang, Dai, Wang, Wang, and Davis]{zhang2019man}
Da Zhang, Xiyang Dai, Xin Wang, Yuan-Fang Wang, and Larry~S Davis.
\newblock Man: Moment alignment network for natural language moment retrieval via iterative graph adjustment.
\newblock In \emph{CVPR}, pages 1247--1257, 2019{\natexlab{a}}.

\bibitem[Zhang et~al.(2020{\natexlab{a}})Zhang, Sun, Jing, and Zhou]{DBLP:conf/acl/ZhangSJZ20}
Hao Zhang, Aixin Sun, Wei Jing, and Joey~Tianyi Zhou.
\newblock Span-based localizing network for natural language video localization.
\newblock In \emph{ACL}, pages 6543--6554. Association for Computational Linguistics, 2020{\natexlab{a}}.

\bibitem[Zhang et~al.(2021)Zhang, Sun, Jing, Zhen, Zhou, and Goh]{zhang2021natural}
Hao Zhang, Aixin Sun, Wei Jing, Liangli Zhen, Joey~Tianyi Zhou, and Rick Siow~Mong Goh.
\newblock Natural language video localization: A revisit in span-based question answering framework.
\newblock \emph{IEEE TPAMI}, 44\penalty0 (8):\penalty0 4252--4266, 2021.

\bibitem[Zhang et~al.(2022{\natexlab{a}})Zhang, Cai, Pan, Hong, Guo, Yang, and Liu]{zhang2022motiondiffuse}
Mingyuan Zhang, Zhongang Cai, Liang Pan, Fangzhou Hong, Xinying Guo, Lei Yang, and Ziwei Liu.
\newblock Motiondiffuse: Text-driven human motion generation with diffusion model.
\newblock \emph{arXiv preprint arXiv:2208.15001}, 2022{\natexlab{a}}.

\bibitem[Zhang et~al.(2020{\natexlab{b}})Zhang, Peng, Fu, and Luo]{zhang2020learning}
Songyang Zhang, Houwen Peng, Jianlong Fu, and Jiebo Luo.
\newblock Learning 2d temporal adjacent networks for moment localization with natural language.
\newblock In \emph{AAAI}, pages 12870--12877, 2020{\natexlab{b}}.

\bibitem[Zhang et~al.(2022{\natexlab{b}})Zhang, Peng, Fu, Lu, and Luo]{DBLP:journals/pami/ZhangPFLL22}
Songyang Zhang, Houwen Peng, Jianlong Fu, Yijuan Lu, and Jiebo Luo.
\newblock Multi-scale 2d temporal adjacency networks for moment localization with natural language.
\newblock \emph{IEEE TPAMI}, 44\penalty0 (12):\penalty0 9073--9087, 2022{\natexlab{b}}.

\bibitem[Zhang et~al.(2019{\natexlab{b}})Zhang, Lin, Zhao, and Xiao]{DBLP:conf/sigir/ZhangLZX19}
Zhu Zhang, Zhijie Lin, Zhou Zhao, and Zhenxin Xiao.
\newblock Cross-modal interaction networks for query-based moment retrieval in videos.
\newblock In \emph{SIGIR}, pages 655--664. {ACM}, 2019{\natexlab{b}}.

\bibitem[Zhao et~al.(2023)Zhao, Lin, Yan, and Li]{zhao2023diffusionvmr}
Henghao Zhao, Kevin~Qinghong Lin, Rui Yan, and Zechao Li.
\newblock Diffusionvmr: Diffusion model for video moment retrieval, 2023.

\end{thebibliography}
}


\end{document}